**[Title Page]**

**An Ensemble Learning Based Approach to Multi-label Power Text Classification for Fault-type Recognition**

**Xiaona Chen, Tanvir Ahmad, Yinglong Ma***

**School of Control and Computer Engineering, North China Electric Power University, Beijing 102206, China.**

**Email: 2920552929@qq.com ;tanvirahmad@ncepu.edu.cn; yinglongma@ncepu.edu.cn.**

**\* Correspondence author**

**Correspondence information:**

**Full name: Yinglong Ma**

**Affiliation: School of Control and Computer Engineering, North China Electric Power University, Beijing 102206, China,**

**Email address: yinglongma@ncepu.edu.cn**

**Telephone number: +86 10 61772643**

# An Ensemble Learning Based Approach to Multi-label Power Text Classification for Fault-type Recognition

Xiaona Chen, Tanvir Ahmad, Yinglong Ma[*]

*School of Control and Computer Engineering, North China Electric Power University, Beijing 102206, China*

**Abstract**

With the rapid development of ICT Custom Services (ICT CS) in power industries, the deployed power ICT CS systems mainly rely on the experience of customer service staff for fault type recognition, questioning, and answering, which makes it difficult and inefficient to precisely resolve the problems issued by users. To resolve this problem, in this paper, firstly, a multi-label fault text classification ensemble approach called BR-GBDT is proposed by combining Binary Relevance and Gradient Boosting Decision Tree for assisted fault type diagnosis and improving the accuracy of fault type recognition. Second, for the problem that there is lack of the training set for power ICT multi-label text classification, an automatic approach is presented to construct the training set from the historical fault text data stored in power ICT CS systems. The extensive experiments were made based on the power ICT CS training set and some general-purpose benchmark training datasets. The experiment results show that our approach outperforms the well known ensemble learning based approaches BR+LR and ML-KNN for fault text classification, efficiently handling the multi-label classification of ICT custom service text data for fault type recognition.

[*] Correspondence author. Tel:+86 10 6177264, Email address: yinglongma@ncepu.edu.cn

## 1. Introduction

The electric power information communication customer service systems (ICT CS systems) play a critical role in the day-to-day processing and business operations as the necessary support for the operation and management of the electrical power companies. The ICT CS systems are used for providing online consultation services for internal and external users of the electrical power companies. According to the statistics of state gride cooperation of china (SGCC) to date, 73 power ICT information systems have been incorporated into the Customer Service Handling System headquarters in SGCC and 24 have been deployed by provincial companies covering many sectors, including human resources, finance, materials, and technical applications. The key task of the ICT customer service system is to examine the specific fault occurrences posed by ICT application users, by identifying all possible triggers and promptly feeding them back to the users. Customer service currently manages professional issues manually through the telephone, which relies on personal technical skills and experience of the handler. With the increasing number of the deployed ICT systems, they are becoming more and more complex. However, due to the limitations of human resources in customer services, it has become costly and difficult for customer services to address the problems efficiently and timely. Online ICT CS systems are ur gently needed to be developed to boost up online corporate processing efficiency by mining historical customer service data.

The rapid development of artificial intelligence and natural language processing technology enables automatic and intelligent analysis based on power ICT CS systems. Chinese historical data stored in power ICT CS systems can be deeply exploited by some machine learning approaches for identifying the fault information about the ICT system such as fault types and causes of such faults, which is the premise towards intelligent questioning and answering of ICT CS systems.

In the last decades, we have witnessed the rapid applications of machine learning classification techniques in many application fields including electric power systems, such as Support Vector Machines (SVM) [1,2,3], Hidden Markov Model (HMM) [5], Artificial Neural Networks (ANN) [6], Bidirectional Long Short-Term Memory (BiLSTM)[8], K-Nearest Neighbor(KNN) [9] and Bayesian network [11], Decision Tree [41], etc. Recently, Chinese text classification models have been presented for text data mining in the field of the electric power industry [13]. These text classification models are often binary classification or multi-class classification models that predict the most likely category label for a text instance from two or multiple categories [16].

However, we argue that text classification based on ICT CS text data can be regarded as a typical scenario for multi-label classification 错误!未找到引用源。, which needs to predict several faults category labels for each fault text. The fault text data of ICT CS systems in the power field may cover many business fields such as desktop applications, human resources, finance, assets, and e-commerce, etc. The fault categories about customer service text data are very likely to come from multiple sources such as business management, information communication, computer hardware, and software, etc. For example, in terms of business management, system faults may be caused by incorrect login account passwords or irregular business operations, and therefore they possibly have multiple fault types at the same time. For another example, a page display fault may be caused by network connection problems, browser problems, database problems, and so on. In these complex scenarios, a fault instance occurred in the power ICT CS system usually corresponds to different category labels of fault types from multiple sources, but current multi-class classification models cannot perform the multi-label classification tasks for fault-type recognition in the power ICT CS system.

So far, there is seldom relevant research on the processing of multi-label text classification in the field of electric power with the exception to our previous work [17]. One of the reasons behind is that a multi-label classification problem often is more complicated [18]. First of all, it is uncertain that how many category labels are corresponding to the sample instances, some sample instances may be only correspond to one category while others may be correspond to many or even hundreds. Secondly, there are usually implicit correlations between different category labels. For example, the samples belonging to the category label "monitor interruption" are also belonging to "monitor exception" with a high probability due to the fact that monitor interruption is a kind of monitor exception. The custom service staffs for the Power ICT CS system cannot accurately capture the implicit interdependency and corelation due to the lack of human resources and the limited expertise and training. After all, the expertise for recognizing fault types covers business process, communication, software, and hardware, etc. What is more, there is no the power text data set to train multi-label text classification models for fault-type recognition in power ICT CS systems. Manually building the multi-label training data set is rather time-consuming and labor-intensive, which is especially true for building a power ICT CS training dataset. It requires human to ensure each sample instance could correspond to a subset of the label set accurately. All of these reasons make it challenging for ICT customer services to achieve high-quality multi-label text classification based on power ICT CS Chinese text data.

In this paper, aiming at the multi-label text classification (MTC) based on power ICT CS systems, we propose a multi-label text classification approach based on automation and high precision ensemble learning in order to assist the customer service in recognizing fault categories (fault types). The contributions of this paper are as follows.

First, for the practical problem that the power training set for ICT multi-label classification is lack and also difficult to obtain, an automatic approach is presented to construct the power training set of ICT custom service text to achieve ICT customer service text multi-label classification efficiently. Second, we present a BR-GBDT based ensemble approach for multi-label text classification, which combines Binary Relevance (BR) [20] and Gradient Boosting Decision Tree (GBDT) [22] to improve the classification precision. Since the proposed BR-GBDT approach fully considers the possible interrelationship between different classification labels, it enables to improve the accuracy of multi-label classification effectively. Third, we made extensive experiments based on the power ICT CS training set and some general-purpose benchmark training datasets in order to validate that our approach outperforms the most popular approaches such as BR+LR and ML-KNN for fault text classification, efficiently handling the multi-label classification of ICT custom service text data for fault type recognition.

The rest of the paper is organized as follows: Section 2 describes ICT customer's text data features and automatic construction of dataset from those data. In Section 3 our proposed method is presented. Section 4 describes the experimental setup, results, and discussion. Section 5 is the conclusion and future work.

## 2. Automatic Construction of Multi-label Training Dataset

In this section, we will discuss how to automatically construct a multi-label training data set for the power ICT CS fault texts, which can be used to train the BR-GBDT based multi-label classification model for power ICT CS fault-type recognition. We need to analyze and describe some typical features in the current power ICT CS fault texts, and then present our automatic construction approach of the multi-label training dataset for the power ICT CS fault texts based on these feature analysis.

## 2.1 Features Analysis of ICT CS Fault Text Data

The power ICT customer service text data instances include fault accounting records and work forms for fault handling Current text data instances are represented in a mixture of semi-structured and unstructured fashion. These Chinese text instances usually explain the specific phenomenon description of the occurred faults and the description of how to handle these faults, the job order number of the agent, and so on. All these elements are usually distinguished through tables, where the content in every table cell is present in unstructured text formats. In addition, there is interdependence and corelation between the fault description and fault types. Specifically speaking, the ICT customer service text data typically has the following features.

First, there are significant variations in terminology in reports for the same fault issue because of the big variation in customer service in terms of specific technical expertise and experience. For example, the fault problems with accessing the database may be reported as "database issues," "database anomalies," or "data access error". Faults occurred in the monitoring system can also be reported as "monitoring system abnormalities". In these cases, we will adopt an effecient text representation method based on word embedding technique that makes comparable these words with similar meanings.

Second, there are great variations in detail degrees about basic ICT fault descriptions in the instances of customer service text data. By explaining the fault reason sometimes, it is maybe possible to describe it in a few words, and sometimes it needs to write a full-length paragraph for describing the fault issue and its causes. In the case, the text feature representation based on customer service text data should fully consider both the short and full-length texts. The feature vector space using word embedding text representation should be kept in a reasonable dimension.

Third, fault accounting records and work forms have no standard format. There is also incomplete information in the text instances of ICT customer service data. There may be miss information about some fault occurrences. This means that these fault text instances include noising information, and they need to be cleaned at the stage of data processing.

Four, a very serious problem is that every fault text intance is not assigned its corresponding fault category labels beforehand in these historical customer service fault text data although they include the descriptions of how to handle these faults that possibly contain the information of fault categories. The description of handling a fault consists of several sentences instead of the set of fault category labels. So we need to extract all the fault category labels (fault types) from the description of handling a fault for every fault instance. The extracted fault category labels from a fault text instance will be assigned the corresponding fault category labels of the fault instance.

All of these practical problems in ICT customer service text data make it difficult to constructe Power ICT CS multi-label training dataset for efficient fault type recognition. In the next subsection, we will discuss how to automatically construct a multi-label training dataset from the historical power ICT CS text data.

## 2.2 Automatic Construction Approach

From the analysis in the previous subsection, the automatic construction approach includes three stages: a data pre-processing stage that includes data cleaning and Chinese word segmentation, a text representation stage where all the Chinese words are encoded as low-dimensional vectors, and a fault type labels mining stage where every faut text instance is assigned several labels. .

## 2.2.1 Pre-processing of fault text data

The customer service data is first appropriately cleaned to eliminate noisy data that is not relevant to the classification in instances. For example, the historical power ICT CS text data often includes the job order numbers, the numbers of custom service staffs, names of departments, the work order number, contact person, caller number in work orders, etc. Such information that has little to do with the classification should be removed for improving and speeding up the efficiency of model training based on training dataset in the future since they are not necessary for the classification algorithm, and even are nosing.

The next is for Chinese words segmentation and the stop words removal from the Chinese text. In this paper, the Jieba word segmentation tool [27] is applied to segment the Chinese paragraphs in the work form texts and fault accounting records. Durning word segmentation, we add a dictionary about the technical terms in the field of ICT custom services for preventing a technical term from being divided into multiple unrelated words. The power ICT CS dictionary includes some typical technical terms in the field of power ICT custom services. Typically, a Chinese phrase consists of continuous words without any separators, unlike an English sentence where words are separated by the space character. For example, the phrase "Monitoring system abnormality" in Chinese is "监控系统异常" that is a single technical term and cannot be divided into multiple Chinese words. Finally, some stop words should be removed from the set of segmented words. Here, the stop words refer to some Chinese words very similar to English words such as "the", "is", "at", "which", "on", "who", and "which", and so on. The procedure of ICT CS text pre-proccessing is shown in the Figure 1.

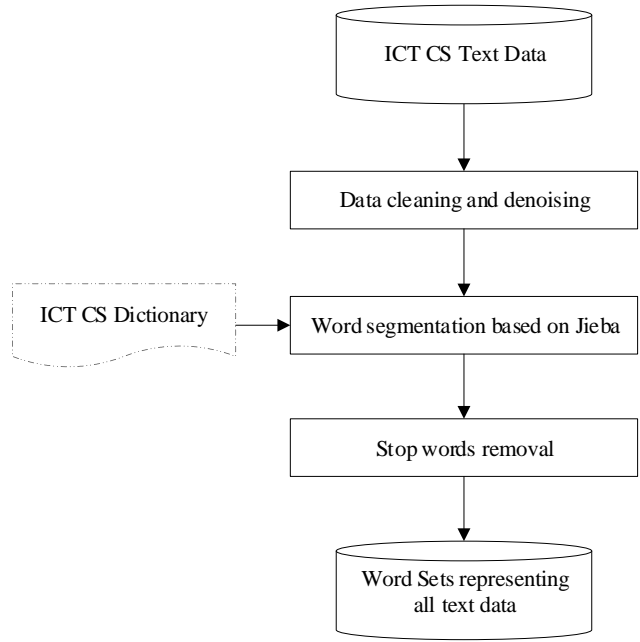

**Fig. 1  Procedure of ICT CS text pre-proccessing**

After a serials of data preprocessing, every fault text instance is converted into a set of Chinese words. All word sets will be used as the text processing corpus for the subsequent processing.

### 2.2.2 Feature representation of fault text instances

Word embedding techniques [28] recently have been widely applied for text representation. In this paper, we use the GloVe word embedding [28] for text presentation, which is based on the continuous bag of words (CBOW) model and converts a text instance into a computer-processable feature semantic vector representation model. This approach is very helpful in dimensionality reduction of corpus word set. The word embedding approach takes the words set as input for the ICT customer service corpus.

A word embedding representation set of all words which appear in the corpus word set is taken as the output via a shallow neural network model. The embedding representation of a word is to use a low-dimensional real number vector to represent the semantic information. The semantic relationship between two words can also be reflected by the distance of the corresponding word embedding vector. And the words with similar semantics usually have closer distance between word vectors. In this paper, the word embedding dimension of each word is set to 400. The larger the embedding dimension, the higher the complexity of model training. Because our data set is domain specific, the dimension 400 is reasonable for our subsequent model training.

Finally, each text instance $d_j$ is represented by a feature representation vector $\vec{d}_j$, which is obtained by calculating the average of summing all word embedding vectors corresponding to each word contained in $d_j$.

### 2.2.3 Fault category label mining for multi-label association

Power ICT customer services do not have open hierarchies of fault classification and vocabulary. Thus, the core feature word sets have to be mined to define the faults and device type features of ICT customer services from historical text data. Term Frequency - Inverse Document Frequency (TF-IDF) model [31] is not only an efficient calculation model based on statistical expertise, but also has great semantic interpretability. TF-IDF model has been widely applied in information retrieval and feature extraction [32], etc. Based on the text processing corpus generated afterword segmentation, we use the TF-IDF model to quantify the importance of each word in the word set. According to the importance, the most important $q$ feature words are selected as classification labels.

Finally, according to the word set and core feature words, the text instances are associated with the corresponding classification labels.

Suppose the frequency of occurrence of the word $w_i$ in text $d_j$ is denoted as $TF_{i,j}$, while the inverse frequency of occurrence of the word $w_i$ in all texts is denoted as $IDF_i$. Mathematically it can be described as follows.

$$TF_{i,j} = n_{i,j} \Big/ \sum_{k=1}^{N} n_{k,j} \tag{1}$$

$$IDF_i = \log(|\{d_j \in D\}| / (|\{j : w_i \in D_j\}| + 1)) \tag{2}$$

$$TF\text{-}IDF_{i,j} = TF_{i,j} \times IDF_i \tag{3}$$

Where $D$ is the corpus word set of text data, $N=|D|$, $D_j$ is the word set of the text $d_j$, and $n_{i,j}$ is the number of occurrences that $w_i$ appears in the text $d_j$. It is easy to find that the larger the value of $TF_{i,j}$, the better the ability of $w_i$ to distinguish features, making it of higher importance of $w_i$ to $D_j$. The larger the $IDF_i$ value, the higher the frequency of occurrence of $w_i$ in the whole texts, which makes $w_i$ more trivial to $D$ for distinguishing between features.

A given text $d_j$ can be associated with all the corresponding classification labels by calculating the value of $TF\text{-}IDF_{i,j}$. Let $|D_j|=m$ be the number of words in the word set $D_j$ of $d_j$. Each word $w_{i,j}$ in $D_j$ will be sorted according to its $TF\text{-}IDF_{i,j}$ value for any $1 \leq p < q \leq m$ and $TF\text{-}IDF_{p,j} \leq TF\text{-}IDF_{q,j}$. Then the set of classification labels associated with $d_j$ is recorded as $L_j \subseteq D_j$, where $L_j$ is the set of classification labels associated with $d_j$, as presented in equation 4.

$$L_j = \{w_{1,j}, ..., w_{k,j} \mid \frac{\sum_{i=1}^{k-1} TF\text{-}IDF_{i,j}}{\sum_{i=1}^{|D_j|} TF\text{-}IDF_{i,j}} < \delta \wedge \frac{\sum_{i=1}^{k} TF\text{-}IDF_{i,j}}{\sum_{i=1}^{|D_j|} TF\text{-}IDF_{i,j}} \geq \delta\} \tag{4}$$

The threshold $\delta$ value is set to 0.5, which can be used to find the least $k$ words that can characterize the text $d_j$ and serve as the associated classification label. The total classification label set of ICT customer service fault text data is denoted as $L$, which is defined in formula 5.

$$L = \bigcup_{i=1}^{k} L_i \tag{5}$$

Finally, we can construct the multi-label training sample set $T$, i.e.,

$$T = \bigcup_{j=1}^{|D|} \{(\vec{d}_j, L_j\} \tag{6}$$

where $\vec{d}_j$ is the embedding vector corresponding to the text $d_j \in D$, and $L_j$ is the set of classification labels associated with $d_j$.

### 2.3 The Constructed ICT CS Multi-label Training Dataset (ICT data)

The ICT customer service fault data from the first half of 2018 are collected as the initial data set according to the multi-label training dataset development approach proposed above. It consists of 1817 instances, including 161 valid fault accounts records and 1656 valid records from office workers. By using data cleaning, word segmentation, word embedding feature representation, multi-label relevance, etc., techniques a multi-label training set as described in section 2, is automatically constructed for the ICT customer service text dataset. As shown in Table 1, a total of 22 labels for the classification are excavated.

**Table 2. Automated constructed ICT fault category labels**

| No. | Category Label Name | No. | Category Label Name |
|---|---|---|---|
| 1 | I6000 interface issues | 12 | Unavailable functions |
| 2 | Monitoring system abnormality | 13 | I6000 operating index abnormality |
| 3 | Excessive server load | 14 | Unavailable users |
| 4 | Optical fibers issues | 15 | Monitoring index abnormal |
| 5 | Platform software issues | 16 | Dispatch data network interruption |
| 6 | Page display issues | 17 | System outage |
| 7 | Device issues | 18 | I6000 monitoring abnormality |
| 8 | Database issues | 19 | Unavailable cascading connections |
| 9 | Network issues | 20 | Health duration interruption |
| 10 | Browser issues | 21 | Locked database users |
| 11 | Host issues | 22 | E-commerce platform issues |

From the training data set, at the same time up to 9 fault categories can be correlated with fault text instances, minimum one and up to 3 or 4 are mostly correlated. We asked experts in the field of electrical power to check and verify the accuracy of the category making on each sample, ensuring that the category marking on the training dataset is accurate. This thoroughly demonstrates that the proposed approach for building the multi-label training set is accurate and generates a high-quality training set. It offers a significant model training support on the ICT customer service framework for automatic fault type with high precision classification.

Our proposed method presented for the construction of an automatic multi-label training set is a general multi-label classification algorithm which not only provides training data for our approach, it can also be used for the training data service for the other data mining on ICT customer service data.

## 3. Proposed Method: BR-GBDT

### 3.1 Problem Statement of Power ICT CS multi-label text classification

Technically speaking, typical methods used for multi-label classification can be roughly divided into two strategies [19]: the Algorithm Adaptation and Problem Transferring. Algorithm Adaptation manages multi-label data directly by transforming and extending the existing learning algorithms. The well-known Algorithm Adaptive approaches corresponding to multi-label classification algorithms are ML-KNN [35], Rank-SVM [36]. However, a more complicated model or higher computational complexity may be caused by expanding the original algorithm model. In contrast, Problem Transfer primarily disassembles the problem of multi-label classification into a multi-classification problem or multiple binary classification problems, such as Binary Relevance (BR) algorithm [20], Classifier Chain (CC) [38].

ICT customer service text data is characterized by high dimensions and sparseness. If the Adaptation Algorithm is adopted, high computational complexity will be caused due to the high dimensionality of the multi-label classification, which will cause additional difficulties for a classification task. Although combining BR with a simple binary classifier is relatively simple, it ignores the semantic relevance between classification labels. Considering that the labels crossing multiple knowledge fields are contained in the ICT customer service, ignoring the semantic relevance between labels will inevitably reduce the accuracy of multi-label classification.

Another problem is that the power ICT CS multi-label text dataset is the typical unbalanced dataset where a fault instance occurred in the power ICT CS system usually corresponds to different category labels of fault types from multiple sources. Recently, some techniques combine single label classifiers to handle multilabel classification, or make modifications to single-label classifiers algorithms to allow their usage for multi-labels classification problems [24]. However, it can be found that many multi-label classifiers perform better when they are applied to balanced datasets, but fail to provide the desired output when implemented in a typical scenario. Therefore, a solution for learning a model from an unbalanced dataset is a need of the day to overcome these issues.,

To solve the above-mentioned problems, we propose the BR-GBDT integrated learning algorithm, which combines the advantages of BR [20] and GBDT [22] for multi-label text classification. Logically this algorithm is relatively simple and easy to expand. A strong classifier is iteratively trained by providing a collection of serialized decision tree weak classifiers to classify different labels and using classification error calculation and negative gradient fitting adjustment of the loss function. The perceptible training process and corrected differences of different labels in the classification result significantly reduce classification bias. Besides, the proposed algorithm also avoids the training overfitting and improves the accuracy of multi-label classification.

### 3.2 Binary Relevance

Probably the most intuitive solution for learning from multi-label examples is binary relevance. It works by breaking down the multi-label learning task into several independent binary learning tasks [20]. Let $L=\{l_1, l_2, \ldots, l_q\}$ be the label set, and $|D|=N$ be the total number of samples in the corpus text set. Based on the training sample set $T$, the corresponding training set $T_j$ as defined in formula 7, is created for the label $l_j$ through Binary Relevance.

$$T_j = \{(\vec{d}_i, l_j) \mid l_j \in L_j \wedge 1 \leq i \leq N\}$$

(7)

Based on that, the strong classifier $p_j : X \rightarrow R$ is trained for each $l_j$ using the GBDT algorithm, where $X$ represents the set of feature vector representations of all texts in $D$, and $R$ is the set of real numbers.

For sample $\vec{d_i}$ with unknown labels, we predict its associated label set $L_i$ by querying each GBDT classifier.

$$L_i = \{ y_j \mid p_j(\vec{d_i}) > 0, 1 \le j \le q \} \tag{8}$$

If all the values of $p_j$ are negative, an empty label set is generated. To avoid the situation of empty prediction, the label with the largest output value is used as the prediction label $\vec{d_i}$.

### 3.3 Gradient Boosting Decision Tree Algorithm

The gradient boosting algorithm is trained in an iterative way. The previous output data is again calculated through negative gradients in the next iteration, previous output errors are corrected by fitting the negative gradient of the mean squarde error (MSE) loss function. Finally, we need to find a fitting function $f_0(x)$ that infinitely close to the true value. The basic principles are as follows. Firstly, we initialize the inaccurate $f_0(x)$ and calculate the residual value $r_o(x) = y - f_o(x)$ based on the difference between $f_0(x)$ and the true value y. Then, to find a suitable $h_o$ to fit $r_o$, the new fitting function in the next round can be represented as $f_{i+1}(x) = f_i(x) + h_i(x)$ and the new residual value is $r_{i+1}(x) = y - f_i(x)$. In this way, iterative correction is made continuously, so that $f(x)$ is continuously approached towards the true value $y$. Gradient boosting algorithm has convergence, which can ensure local or global optimization.

Concretely, we initialize the weak learner as formula 9.

$$f_0(x) = \arg\min \sum_{i=1}^{N} L(y_i, \rho) \tag{9}$$

We determine the constant $\rho$ that can make the initial prediction loss MSE function $L$ obtaining the minimum value. The loss function can be represented as the formula 10.

$$L(y, f(x)) = (y - f(x))^2 \tag{10}$$

For all samples $N$, the total loss is defined as.

$$L_{all} = \sum_{i=1}^{N} L(y_i, f_m(x_i)) \tag{11}$$

The initialization aims to minimize the value of the loss and to find the fastest gradient descent path. The negative gradient function calculated in each iteration is represented in equation 12.

$$-g(x_i) = -\frac{\partial L(y_i, f(x_i))}{\partial f(x_i)} \tag{12}$$

To fit the negative gradient $-g(x_i)$, a fitting function $h(x_i; \alpha)$ is defined.

$$\alpha_m = \arg\min \sum_{i=1}^{N}(-g(x_i) - \beta h(x_i; \alpha))^2 \tag{13}$$

We can calculate the optimal value $\beta$ according to the following formula.

$$\beta_m = \arg\min \sum_{i=1}^{N} L(y_i, f_{m-1}(x_i) + \beta h(x_i; \alpha_m)) \tag{14}$$

Finally, the calculation results are merged into the model, updating the prediction function, iteration is terminated upon reaching the pre-defined number of iterations. The final step is Mathematically expressed as.

$$f_m(x) = f_{m-1}(x) + \beta_m h_m(x; a_m) \tag{15}$$

### 3.4 Algorithm for BR-GBDT Based Multi-label Classification

By combining Binary Relevance model and the Gradient Boosting Decision Tree model, we designed the BR-GBDT algorithm for multi-label text classification based on the power ICT CS text data. The pseudocode of BR-GBDT model is presented in Algorithm 1.

---

**Algorithm 1**: MTC based on BR-GBDT for customer service Chinese text

---

**Input**: (1) ICT dataset $T = \{(\vec{d}_j, L_j)\}_{j=1}^{N}$ with $N$ samples, where $\vec{d}_j$ is the text feature vector, and $L_i$ is the label vector set corresponding to $\vec{d}_j$. (2) $M$: the number of the iteration. (3) $q$: the total number of labels. (4) $\vec{d}$: unprocessed samples.

**Output**: The set of prediction labels $Y$ on $\vec{d}$

Begin

1.  **for** $j$=1 to $q$ do

2.      Create a binary data set $T_j = \{(\vec{d}_i, l_j) \mid l_j \in L_j \wedge 1 \leq i \leq N\}$

---

3.  Initialize the weak classifier $f_0(x) = \arg\min \sum_{i=1}^{N} L(l_i, \rho)$

4.  **for** $m = 1$ to $M$ do

5.  Calculate the negative gradient $-g(\vec{d}_i) = -\dfrac{\partial L(l_i, f(\vec{d}_i))}{\partial f(\vec{d}_i)}$, $i = 1, 2, ..., N$

6.  Fit new function $h(\vec{d}_i; \alpha)$

7.  Calculate the parameters $\alpha_m = \arg\min \sum_{i=1}^{N} (-g_m(\vec{d}_i) - \beta h(\vec{d}_i; \alpha))^2$

8.  Weighting factor $\beta_m = \arg\min \sum_{i=1}^{N} L(l_i, f_{m-1}(\vec{d}_i) + \beta h(\vec{d}_i; \alpha_m))$

9.  Update function $f_m(\vec{d}) = f_{m-1}(\vec{d}) + \beta_m h_m(\vec{d}; \alpha_m)$

10. **endfor**

11. Derive the classifier function $p_j \leftarrow f(D_j)$.

12. **endfor**

13. $Y = \{l_j \mid p_j(\vec{d}) > 0, 1 \le j \le q\}$

14. if $Y = \varnothing$ then $Y = \{l_{j*} \mid j* = \arg\max_{1 \le j \le q} p_j(\vec{d})\}$

15. **return** $Y$

End

## 4. Experimental Evaluationp and Analysis

The classification task essentially aims at discovering information. This can be used to predict an unknown instance class, based on attribute values that describe such an instance. As a result, we can divide the classification tasks by the number of labels for each category to be predicted. All the experiments are conducted using the Ubuntu 16.04 operating system having Intel Core i3-4130 CPU, using Python programming language for implantation of the algorithm.

### 4.1 Datasets for Training and Testing

For training and testing of our proposed BR-GBDT model, we used three datasets. One is our power ICT CS multi-label traning dataset (ICT Data) described in the previous section in detail. The other two datasets are from [39]: TMC2007 and RCV1-v2. The two datasets are often used as the general-purpose benckmark datasets for multi-label text classification. We split the datasets into training and testing sets, 80% of the data from each of the above data sets is randomly selected as the training set for model training, and 20% for the testing. Table 2 presents details about the number of samples and categories of each dataset.

**Table 2. Information of data sets**

| Name | Form | Sample Number | Category Number |
|------|------|---------------|-----------------|
| TMC2007 | Text | 28596 | 22 |
| RCV1-v2 | Text | 6000 | 101 |
| ICT data | Text | 1817 | 22 |

### 4.2 Performance Indexes

The performance of the algorithm is evaluated by the classical multi-label classification performance index, which are accuracy rate, recall rate, and F1 value. To evaluate the indices, the following four data must be counted when calculating the performance of the prediction of a multilabel classification model.

(1) Precision: $P = \dfrac{TP}{TP + FP}$

(2) Recall: $R = \dfrac{TP}{TP + FN}$

(3) F1: $F1 = \dfrac{2PR}{P + R}$

Where True positive (TP) represents the number of instances predicting positive class as positive, False Positive (FP) represents the number of instances predicting negative class as positive, and False Negative (FN) is the number of predicting positive class as the negative class.

### 4.3 Baselines

We evaluate our proposed BR-GBDT model by comparing the performance with two well-known ensemble learning based multi-label classification models:BR+LR and ML-KNN. Both the approaches, respectively representing two kinds of ensemble learning methods including Problem Transferring and Algorithm Adaptation, are used as the algorithm baselines to verify the effectiveness of our proposed BR-GBDT method. The BR+LR approach refers to combining the Binary Relevance with the Logical Regression method, which is a typical Problem Transferring based ensemble learning approach. ML-KNN is a well-known Algorithm Adaptation based ensemble learning method.

## 4.4 Results and Analysis

We evaluated our BR-GBDT approach over the three benchmark datasets in comparison with the baseline algorithms BR+LR and ML-KNN. To analyze the experimental results, the three performance indexes including precision rate, recall rate, and F1 were used. Experimental results are presented in Tables 3, 4, and 5, respectively.

From Table 3, it is worth noting that our proposed method BR-GBDT has got significantly better performance than BR+LR, ML-KNN, algorithms in terms of precision. By comparing the recall rate results in Table 4, in terms of the recall rate index, BR-GBDT has achieved the best average rank than BR+LR and ML-KNN. BR-GBDT has an accuracy rate of more than 90%, especially on the ICT dataset. This shows that the proposed method is better and reliable than other classical multi-label classification algorithms to be used for assisting the ICT customer service system for fault type recognition.

Comprehensive classification performance measured by the F1 score is presented in Table 5. One can clearly note that the proposed BR-GBDT method had better performance than BR+LR and ML-KNN algorithm. From the experimental results, it is clear that BR-GBDT has better classification prediction capabilities on large scale as well on small scale datasets. This strong versatility makes it useful for model training in various text classification models or other AI, machine learning models such as clustering algorithms. This shows that the proposed BR-GBDT is an efficient approach for fault type assessment on ICT customer service systems, which is of great significance to boost the online consulting service efficiency.

### Table 3. Precision comparison

|         | BR+LR  | ML-kNN | BR-GBDT |
|---------|--------|--------|---------|
| TMC2007 | 0.6252 | 0.6072 | 0.6650  |
| RCV1-v2 | 0.8663 | 0.7808 | 0.8361  |
| ICTdata | 0.7660 | 0.8611 | 0.9013  |

**Table 4. Recall comparison**

|          | BR+LR  | ML-kNN | BR-GBDT |
|----------|--------|--------|---------|
| TMC2007  | 0.5586 | 0.6171 | 0.6520  |
| RCV1-v2  | 0.4250 | 0.7064 | 0.7280  |
| ICTdata  | 0.7061 | 0.8316 | 0.8743  |

**Table 5. F1 comparison**

|          | BR+LR  | ML-kNN | BR-GBDT |
|----------|--------|--------|---------|
| TMC2007  | 0.5900 | 0.6121 | 0.6584  |
| RCV1-v2  | 0.5702 | 0.7417 | 0.7783  |
| ICTdata  | 0.7348 | 0.8461 | 0.8735  |

## 5. Conclusion

In this paper, through the in-depth analysis, it was found that the text classification task of ICT customer service text data has a typical multi-label classification application scenario. We analyzed and described the typical features of power ICT CS text data, and further proposed an automatic construction approach to building multi-label training dataset from the historical power ICT CS text data. Considering the characteristics and problems of mluti-label text classification based on powr ICT CS text data, a BR-GBDT-based ensemble learning approach is therefore proposed for multi-label text classification of ICT customer services. The experimental results show that the proposed method can efficiently process the multi-label classification task of ICT customer service text data and outperforms the most popular approaches.

In the future work, we will concentrate on the semantic correlation between category labels in ICT customer service data to improve classification performance. And also, some complex neural network models based multi-label text classification will be explored for further improving the performance of multi-label text classification based on the power ICT customer service text data.

**Acknowledgements**

This work is partially supported by the National Key R&D Program of China under Grant (2018YFC0831404, 2018YFC0830605), and the State Grid Corp of China Science and Technology Project (SGJSDK00KFJS1800443).

**Declarations**

The authors declare that they have no known competing financial interests or personal relationships that could have appeared to influence the work reported in this article. They have no conflicts of interest to declare that are relevant to the content of this article.